\documentclass{article}
\usepackage{graphicx, booktabs, tabularx, threeparttable, amsmath, multirow, subcaption, hyperref}
\usepackage[title]{appendix}
\usepackage{xcolor}
\usepackage{float}

\usepackage[preprint]{corl_2026} 

\title{Getting Out and Getting Back: World and Behavior Grounding in Real2Sim2Real Co-Training}

\author{
  Samuel Liu$^{1,2,*}$, Youngsun Kim$^{2}$, Martin Matak$^{2}$, Gilwoo Lee$^{2}$ \\
  $^{1}$University of Cambridge \quad $^{2}$Industrial Next \\
  $^{*}$Corresponding author: \texttt{syl63@cam.ac.uk}
}

\begin{document}
\maketitle


\begin{abstract}
    Simulation can expand scarce real demonstrations for co-training, yet how world fidelity and similarity to human behavior affect policy performance remains unclear. We distinguish \textbf{world grounding}, which aligns simulation with the real system, and \textbf{behavior grounding}, which aligns simulated trajectories with human motion. We build a real2sim2real pipeline that varies these axes independently to generate data for co-training. On a dynamic dexterous pick-and-sort task, fully grounded co-training raises success from 52\% to 86\%; averaged across configurations, world grounding improves success by 18 percentage points and behavior grounding by 10. Deployed policies behave like a \textit{mixture of real-derived and simulation-derived policies}, imitating real demonstrations in covered states and relying on simulated behavior elsewhere, which we examine through latent-space analysis. Together, these results suggest complementary roles: world grounding lets policies use simulated experience beyond real-data coverage, while behavior grounding matters mainly when world grounding is imperfect. Grounded simulation remains beneficial when co-training foundation models. Our project website is available at: \url{https://industrialnext.github.io/r2s2r-grounding/}.

\end{abstract}



\section{Introduction}
    Human demonstrations, particularly teleoperation, are a primary source of data for robot learning \citep{arxiv241024164,Zhao-RSS-23}. This is especially valuable for dexterous manipulation, where rewards are hard to design and high-dimensional space makes exploration difficult for reinforcement learning \citep{arxiv170910087}. However, collecting diverse and high-quality demonstrations is costly \citep{arxiv250706219,arxiv210803298,dreammachines2026pi05,arxiv231017596,arxiv240312945}, while variations across demonstrations can introduce inconsistency~\citep{arxiv210803298,Zhao-RSS-23}. 

    Simulation offers a way to achieve both diversity and quality. From a small set of real demonstrations, simulation data can be generated at scale, with domain randomization adding diversity \citep{arxiv170306907,arxiv231017596} while success filtering preserves quality \citep{arxiv231017596}. Such data have been shown to be especially valuable when co-trained alongside real data, even when real data is scarce \citep{arxiv250322634,arxiv250324361}.

    However, simulated and real data inevitably differ. To better understand the downstream effect of the sim2real gap, we separate the grounding of the simulation data into two axes: \textbf{World Grounding} and \textbf{Behavior Grounding}. The world gap lies in how the simulator's rendering, physics, and robot control differ from the real system, while the behavior gap lies in how simulated trajectories differ from human demonstrations. Existing real2sim2real pipelines ground simulation in real environments and demonstrations through environment reconstruction, motion refinement, and demonstration adaptation \citep{arxiv240303949, arxiv260608828, arxiv231017596}. Although these approaches improve policy performance, the individual and combined contributions of world fidelity and human-like behavior remain unclear. 

    We hypothesize that world grounding is the primary factor in co-trained policy performance since it determines whether actions learned in simulation have the same effect on the real robot. However, because world grounding is never perfect in practice, behavior grounding plays an important role in keeping simulated behavior close to real demonstrations, which the policy can fall back on.
    
    We investigate the impact of world grounding and behavior grounding on policy performance on a dynamic dexterous pick-and-sort task with objects on a moving conveyor belt. We build a real2sim2real pipeline in which each axis can be independently set to grounded or ungrounded, letting us conduct a 2×2 factorial experiment. Policies for all four configurations are co-trained on 100 real demonstrations and 1,500 simulated episodes and evaluated over 50 real-world rollouts that include both nominal and shifted conditions.

Our main contributions are the following:
\begin{itemize}
    \setlength{\itemsep}{1pt}
    \setlength{\parsep}{1pt}
    \setlength{\parskip}{1pt}
    \item \textbf{Independent grounding control.} We introduce a pipeline that separates world and behavior grounding, enabling a controlled study of their individual and combined effect on dynamic dexterous manipulation.

    \item \textbf{Transition between real-derived and simulation-derived behavior.} We observe that co-trained policies switch between the two within a rollout, adopting simulated behavior in states covered only by simulation, even when it differs substantially from human behavior.

    \item \textbf{Relevance to foundation models.} We show that grounded simulation still benefits post-trained foundation models, extending these findings beyond policies trained from scratch.
\end{itemize}

\begin{figure}[t]
    \centering
    \includegraphics[width=0.95\linewidth]{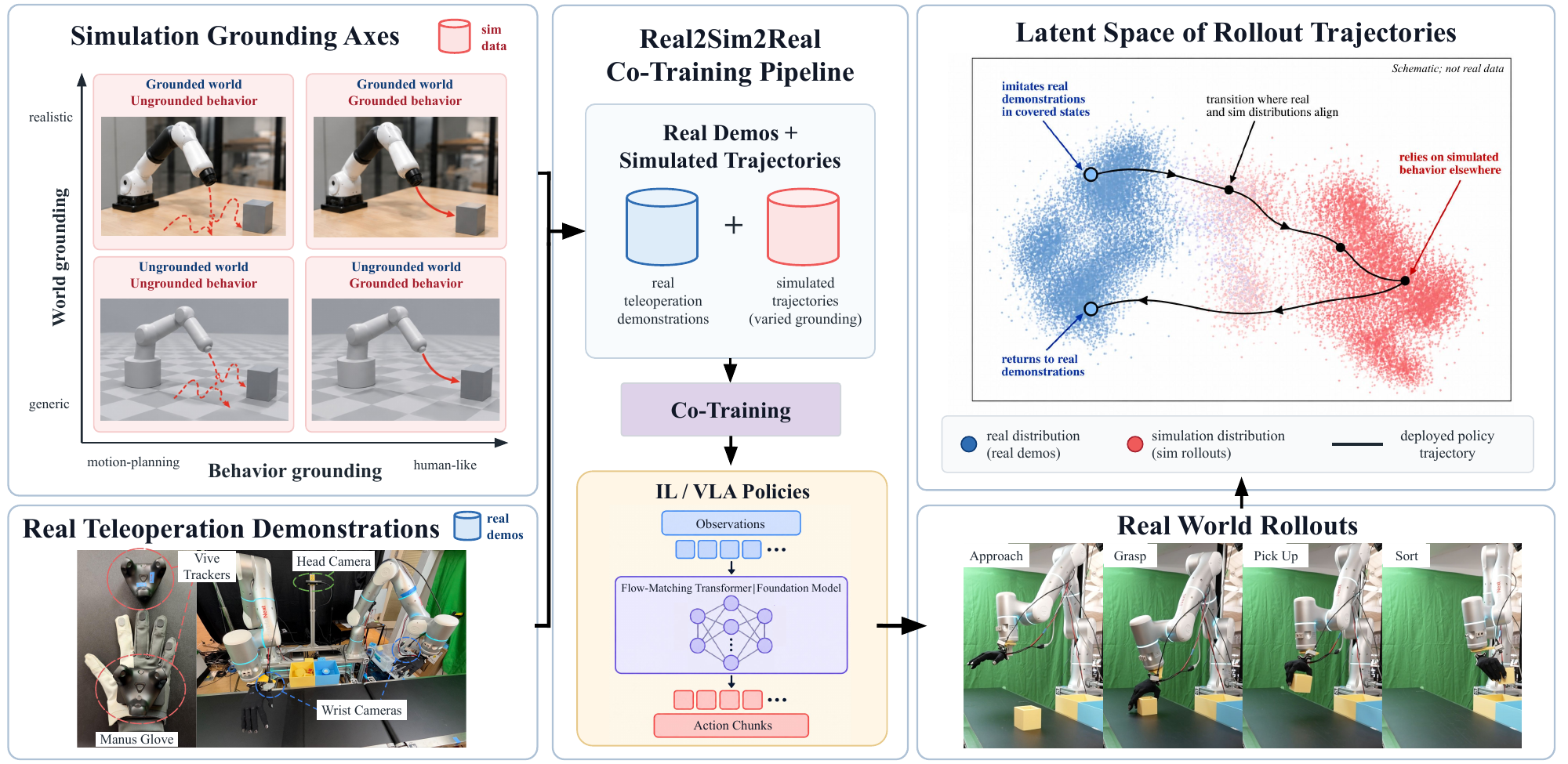}
    \caption{Overview of world and behavior grounding in real2sim2real co-training. (Left) We vary simulated data along two axes, world grounding and behavior grounding, giving four simulated datasets. (Center) Each is combined with 100 real teleoperation demonstrations to co-train policies from scratch or post-train foundation models. (Right) Hypothesized mechanism (schematic, not real data): a deployed policy imitates the real demonstrations in states they cover, relies on simulated behavior elsewhere, and returns to real-like states once it can.}
    \label{fig:overview}
\end{figure}

\section{Related Work}
\label{sec:litreview}

\textbf{Sim-and-Real Co-Training.} Co-training combines scarce real demonstrations with larger simulated datasets, improving generalization to conditions absent from the real data \citep{arxiv250324361,arxiv250322634,arxiv260201067,arxiv260212628}. The benefit depends on simulation mismatch: Wei et al. find a greater sensitivity to physics gaps than to visual gaps, with small visual differences sometimes improving performance \citep{arxiv250322634}. These findings motivate our controlled comparison of world fidelity and trajectory behavior, varying each independently on the same task.

\textbf{World Grounding.} Prior work aligns simulation through camera calibration, scene reconstruction, and identification of robot dynamics and control \citep{arxiv251115200,arxiv240303949,arxiv180410332}, while randomizing appearance to broaden coverage \citep{arxiv170306907}. The importance of each component varies across settings: physical mismatch dominates some co-training results \citep{arxiv250322634}, whereas rendering fidelity matters more than contact parameters in a study of zero-shot VLA transfer \citep{arxiv260322876}. We examine how world grounding affects co-trained policies under a common task and its interaction with behavior grounding.

\textbf{Behavior Grounding.} Synthetic trajectories can be generated by adapting human demonstrations \citep{arxiv231017596,arxiv241024185}, reinforcement learning \citep{arxiv250919752,arxiv260530957}, or motion planning with grasp synthesis \citep{arxiv230516309,arxiv250503233}. These approaches differ in how closely their motions follow human demonstrations. Although machine-generated data can support strong policy performance \citep{arxiv250919752,arxiv260530957}, whether preserving human motion improves co-training independently of world fidelity remains unclear. We compare demonstration-derived and machine-generated trajectories in both grounded and ungrounded worlds.

\section{Method}
\label{sec:method}
    \subsection{Data Generation Pipeline}

Our pipeline generates simulated data under four configurations, setting world grounding and behavior grounding independently to grounded or ungrounded. For each configuration, we generate trajectories using MuJoCo Warp 3.12~\citep{mjwarp} for physics simulation and Isaac Sim 6.0.1~\citep{isaacsim} for rendering. Each trajectory is executed under contact physics and admitted only if it picks up the object and places it in the assigned bin. Each of the four datasets contains about 1,500 trajectories, spanning the same objects and belt-speed range (Appendix~\ref{app:behavior}). These trajectories are then combined with real-world demonstrations per co-training mixture ratio, to train an imitation-learning policy from scratch or fine-tune pre-trained foundation models.

\subsection{World and Behavior Grounding}

\textbf{Grounded vs. Ungrounded World.} For Grounded World, we align simulation with the real system through visual calibration, scene reconstruction, and robot system identification.
We calibrate camera geometry and blur, randomize lighting and object color, register robot and conveyor geometry, and use CAD objects with measured masses.
We use factory-calibrated arm kinematics and fit controllers to recorded responses.
For Ungrounded World, we use ideal cameras, tape-measured poses, a supplied scene layout, nominal kinematics, and calculated controller gains.
Both worlds are rendered with Isaac Sim's native RTX renderer, so they differ in calibration and identification rather than rendering method. Appendix~\ref{app:world} quantifies the mismatches.

\textbf{Grounded vs. Ungrounded Behavior.} For Grounded Behavior, we adopt MimicGen~\cite{arxiv231017596} to shift teleoperated demonstrations to new object poses, sizes, and belt speeds while preserving their motion. We initially take real-world demonstrations and adjust initial object poses estimated by SAM 2~\citep{arxiv240800714} and FoundationPose~\citep{arxiv231208344} for stable belt contact. For Ungrounded Behavior, we synthesize trajectories without human motion references. GraspGen-X~\cite{han2026graspgenx}, adapted to the hand's closure geometry, proposes grasps that are refined against object surfaces, and a motion planner connects the approach, interception, carry, and release phases with collision-checked IK routes.


\section{Experiments and Results}
\label{sec:experiments}
    \subsection{Task Setup}
    Our setup consists of two Flexiv Rizon 4S arms with Wuji Hand 2 hands: the right arm picks objects from a moving conveyor belt and sorts them by color, while the left arm is static and provides an additional camera view. Observations come from two wrist-mounted fisheye cameras and a head-mounted stereo camera, and real demonstrations are collected by teleoperation with a Manus glove and Vive tracker (Appendix~\ref{app:setup}). Unless noted otherwise, each policy is evaluated over 50 real-world trials: 24 under nominal conditions and 26 under shifts in lighting, belt speed, novel objects, or their combination, which are absent from the real demonstrations but present in simulation.

    \begin{figure}[t]
    \centering
    \begin{subfigure}[b]{0.49\linewidth}
        \centering
        \includegraphics[width=\linewidth]{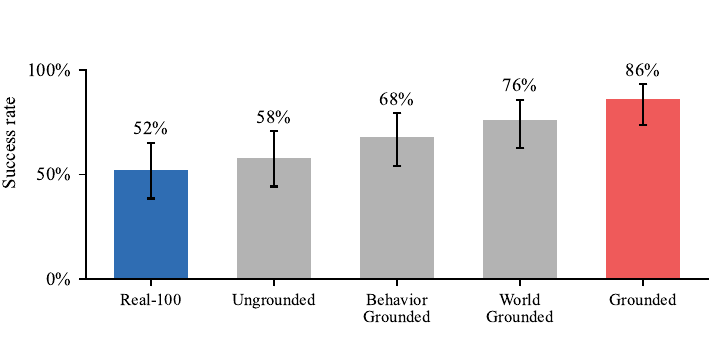}
        \caption{Flow-matching Policy}
        \label{fig:Experiment-1}
    \end{subfigure}
    \hfill
    \begin{subfigure}[b]{0.49\linewidth}
        \centering
        \includegraphics[width=\linewidth]{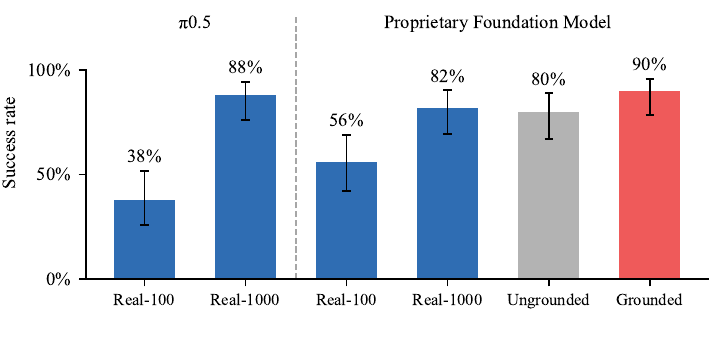}
        \caption{Post-trained Foundation models}
        \label{fig:Experiment-2}
    \end{subfigure}
    \caption{
\textbf{Grounded co-training improves task success,} for both (a) flow-matching policies and (b) foundation models post-trained with sim-real mixture data. All co-trained policies (Ungrounded, Behavior Grounded, World Grounded, Grounded) use 100 real demonstrations plus about 1,500 simulated trajectories. Error bars show 95\% Wilson score intervals over 50 real-world trials.}
    \label{fig:experiments1}
\end{figure}

    \subsection{Real-Only vs. Co-Training}
    We first evaluate whether adding simulated data helps. We train two flow-matching transformer policies: one on 100 real teleoperation demonstrations, and one on the same demonstrations plus about 1,500 world- and behavior-grounded simulated trajectories. Co-training raises success from 52\% to 86\% (Fig.~\ref{fig:Experiment-1}), with the largest gains in conditions the real data does not cover. Simulated data therefore provides a substantial benefit, which motivates asking which aspects of its grounding are responsible.

    \subsection{Effect of World and Behavior Grounding}
    We next compare the four grounding configurations, each co-trained with the same 100 real demonstrations and 1,500 simulated trajectories. Every co-trained policy outperforms the real-only baseline (Fig.~\ref{fig:Experiment-1}). Averaged over the other axis, world grounding improves overall success by 18 percentage points and behavior grounding by 10 percentage points, with world grounding's gains greater in conditions the real data does not cover. Per-condition success counts are given in Appendix~\ref{app:eval}.

    \subsection{Grounding for Foundation Models}
    We also test foundation models, using $\pi_{0.5}$ and our proprietary model (Fig.~\ref{fig:Experiment-2}). With real data alone, going from 100 to 1,000 demonstrations improves success by 50 percentage points for $\pi_{0.5}$ and 26 for ours, so real data remains a bottleneck even with pre-training. Co-training eases it: with 100 real demonstrations and 1,500 fully grounded simulated trajectories, our model reaches 90\%, at least matching 1,000 real demonstrations. Grounding still helps, though less than for the policy trained from scratch (10 vs.\ 28 points over fully ungrounded data), suggesting that pre-training partly compensates for ungrounded data.

        \begin{figure}[t]
    \centering
    \begin{subfigure}[t]{0.30\linewidth}
        \centering
        \includegraphics[width=\linewidth]{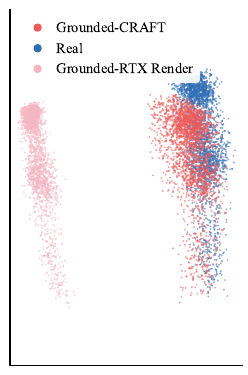}
        \caption{Effect of visual grounding.}        \label{fig:Experiment-3}
    \end{subfigure}
    \hfill
    \begin{subfigure}[t]{0.68\linewidth}
        \centering
        \includegraphics[width=\linewidth]{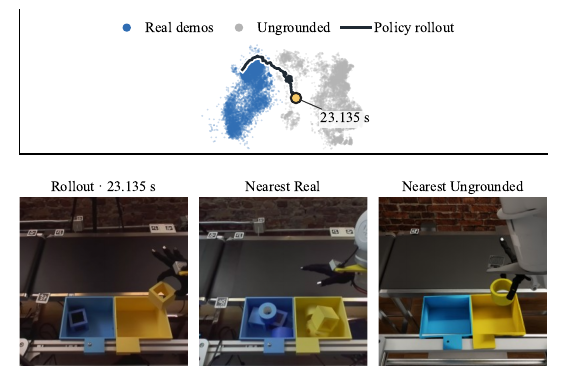}
        \caption{Diagnostic rollout and nearest demonstration frames.}        \label{fig:Experiment-4}
    \end{subfigure}
    \caption{\textbf{Visual grounding and rollout behavior in observation-embedding PCA projections.} (a) CRAFT-style visual transfer places simulated observations closer to real data than native RTX rendering in this projection. (b) A selected rollout of a diagnostic policy trained with 10 real demonstrations and fully ungrounded simulation data moves toward the simulated data region while exhibiting a two-finger pinch, a characteristic of ungrounded simulated trajectories.}
    \label{fig:experiments2}
\end{figure}

    \subsection{Qualitative Observations and Discussion}
Because many of the quantitative comparisons above have overlapping 95\% Wilson score intervals, their differences are best interpreted as trends rather than definitive gaps. We therefore complement them with qualitative observations of how the policies succeed and fail, which help explain these trends and suggest the mechanisms behind them.

\textbf{Switching between real-like and sim-like behaviors.} Similar to observations by~\citet{arxiv250322634}, policies often changed behavior within a single rollout. In a representative rollout of the policy co-trained with fully ungrounded simulation data, the robot began with a real-like approach but missed the object, then switched to visibly sim-like motion, accelerating backwards along the belt to catch the object and complete the pick. This correction likely came from the simulated data: failed teleoperation attempts were discarded, so the real data contains few such recoveries; and the motion was faster and more direct, as is characteristic of our motion-planned trajectories. Co-trained policies thus behave like a mixture of real-derived and simulation-derived behavior, following the real demonstrations in states they cover and relying on simulated behavior elsewhere.

\textbf{Impact of world grounding.} Whenever a policy relies on simulated behavior on the real robot, that behavior succeeds only if simulated actions have the same effect in the real world. This is how co-training generalizes: in conditions the real data does not cover, such as faster belt speeds and new lateral positions, performance depends on how closely the simulated world matches the real one.

\textbf{Impact of behavior grounding.} After the grasp, the robot should return to states resembling the real demonstrations, and in many rollouts it went on to sort the object as expected. In some rollouts, however, it failed to release the object or drifted and missed the bin, suggesting it was still following simulation-derived behavior that the remaining sim-to-real gap made unreliable. These failures occurred mainly in policies trained on world- and behavior-ungrounded data, which have neither accurate simulated dynamics nor human-like simulated behavior to fall back on. We examine these transitions in latent space in Section~\ref{sec:latent}.

\subsection{Latent-Space Analysis}\label{sec:latent}
\textbf{Making transitions apparent.} Rollouts in the main conditions rarely entered the region occupied by simulated data, so to make switching easier to observe, we trained a diagnostic policy with only 10 real demonstrations and 1,500 fully ungrounded simulation data. Fig.~\ref{fig:Experiment-4} shows a selected rollout. Its observations leave the real region and enter a sparsely populated part of the simulated distribution, and its behavior changes accordingly: it grasps the object with a two-finger pinch and moves the arm as in the motion-planned simulation data. The rollout does not return to the real region, and the policy never releases the object, consistent with our hypothesis that, without grounding, a policy that enters simulated behavior cannot find its way back to the real data.

\textbf{Improved world grounding.} Sufficiently accurate world grounding should reduce the need for behavior grounding. In Fig.~\ref{fig:Experiment-1}, however, behavior grounding improved performance by a similar amount under both world-grounding configurations, suggesting that our calibration-based world grounding was not yet accurate enough. We therefore improved its visual fidelity with CRAFT-inspired video translation~\citep{chen2026craft}, after which simulated observations overlap the real ones in latent space, while the RTX-rendered observations form a separate cluster (Fig.~\ref{fig:Experiment-3}). Co-trained with 25 real and 75 simulated trajectories, policies using behavior-grounded and behavior-ungrounded data both succeeded in 17 of 24 trials, on par with our policy trained on 100 real trajectories (16/24). This is consistent with our hypothesis that improved world grounding reduces the need for behavior grounding.

\textbf{Limitations.} This analysis is qualitative and rests on selected rollouts, including a diagnostic policy trained with far less real data than our main conditions. Because rollouts always take place in the real world, their observations rarely fall in the simulated region, and the embedding likely reflects the visual domain of the images as well as the state of the task. It is also hard to tell whether a given motion draws on real or simulated data: distance-based attribution in action space was unreliable, sometimes placing rollouts of a real-only policy closer to simulated actions, and a two-dimensional projection can hide structure in the full embedding space. We therefore present these observations as illustrations of the proposed mechanism rather than direct evidence for it.



\section{Conclusion}
\label{sec:conclusion}
    We separated simulation grounding into two axes, world grounding and behavior grounding, and varied each independently in real2sim2real co-training for dynamic dexterous pick-and-sort. Co-training with fully grounded simulation raised success, and world grounding tended to contribute more than behavior grounding. With a pre-trained foundation model, 100 real demonstrations plus grounded simulation matched fine-tuning on 1,000 real demonstrations. Rollouts suggest that co-trained policies follow real-derived behavior where demonstrations cover the state and switch to simulation-derived behavior elsewhere. This is consistent with our hypothesis that world grounding determines whether this borrowed behavior works on the real robot, and behavior grounding whether the policy can return to demonstrated behavior.

    \subsection{Limitations}
    Our study covers a single task with a limited number of trials per condition, and many differences fall within overlapping confidence intervals. Our grounded world varies visual and dynamic fidelity together, so their individual contributions remain unseparated. The latent-space analysis illustrates rather than establishes the proposed mechanism; attributing actions directly to real or simulated data, and testing longer-horizon tasks where returning to demonstrated behavior should matter more, are directions for future work.


\clearpage
\acknowledgments{This work was conducted at Industrial Next, which provided the robot hardware, compute, and engineering support for this project. We thank Edward Jung for assistance with teleoperation data collection. We also thank Palatial for providing the simulation scene layout.}


\bibliography{paper}  

\clearpage
\begin{appendices}
\section{World Grounding Details}
\label{app:world}

\begin{table}[H]
\centering
\small
\caption{World configurations. All other listed controls are shared.}
\label{tab:world_config}
\begin{tabularx}{\linewidth}{@{}lXX@{}}
\toprule
Factor & Grounded & Ungrounded \\
\midrule
Wrist cameras & Fitted fisheye/mount; blur $\sigma=0.53$ px & Ideal $160^\circ$ fisheye; CAD mount; no blur \\
Head/base/belt poses & Registered using robot motion and AprilTags & Tape-measured head; nominal base orientation; supplied scene layout \\
Arm model/control & Factory kinematics; weighted IK, filtering, nullspace; 29 fitted parameters & Nominal kinematics; damped least-squares IK; calculated gains ($\zeta=0.7$), no armature \\
Hand response & Fitted delay, deadband, and lag & Vendor defaults; direct commands \\
\bottomrule
\end{tabularx}
\end{table}

\textbf{Shared controls.} Object geometry/mass, contact parameters, appearance randomization, initial object poses, belt speed, head intrinsics, hand/TCP registration, and left-arm control are fixed across world conditions. Belt height varies over the same 8~mm range around each world's layout. Contact settings are effective simulation parameters, not measured material constants.

\textbf{Camera and scene validation.} Joint calibration from 14 checkerboard images yields 0.85~px leave-one-out reprojection error versus 6.44~px for the CAD mount with fitted intrinsics. Grounded/ungrounded wrist focal lengths are 252/229~px. Head placement differs by 30.1~mm and $1.7^\circ$; the ungrounded belt is displaced horizontally by 24/18~mm and is 14~mm lower. The stationary left wrist reuses the fitted lens/blur with its CAD mount. External RGB-D reconstruction uses SAM~2 and FoundationPose, accepting initial object poses at held-out mask IoU $\geq0.6$ after stable-contact correction.

\textbf{Robot validation.} The surrogate arm controller is fitted on ten 1~s windows. Continuous free-space replays use a single reset, no measured-state feedback, a fixed hand, and 1~ms steps. TCP position/orientation RMS errors are 1.65~mm/$0.38^\circ$ on a held-out 15~s recording and 12.4~mm/$1.67^\circ$ over the 79~s fitting recording, showing residual long-horizon drift. An earlier fitted controller achieved 1.77~mm versus 2.95~mm for calculated gains; this comparison shares calibrated kinematics and hand response and does not validate the final world configurations. The 20-joint hand model uses 12 training and 6 validation episodes, with median fingertip RMS error 0.64~mm.

\section{Behavior Grounding Details}
\label{app:behavior}
MimicGen transforms demonstration phases to new object poses, sizes, and belt speeds; transformed commands execute through the simulated controller. Ungrounded trajectories use GraspGen-X proposals adapted to hand closure geometry, surface refinement, and collision-checked IK routes for approach, interception, carry, and release. Both conditions retain only successful physics-based pickups and placements into the assigned bin, yielding about 1,500 trajectories per world over the same objects and belt-speed range.

\section{Policy Models}
\label{app:models}
Both models condition on three $256\times256$ RGB views from the head and wrist cameras, 20 proprioceptive samples at 50\,Hz, a text instruction, and an embodiment identifier. A shared pretrained DINOv3 ViT-B/16 image encoder uses frozen backbone weights with rank-8 low-rank adapters in the final four blocks, while a frozen CLIP text encoder and learned resampler provide eight task tokens. The 500M model's policy transformer and action head are trained from scratch on task-specific data; the 2B foundation model is pretrained on robot demonstrations across multiple tasks and embodiments, then fine-tuned for the target task. Both use conditional flow matching with a two-step Euler solver ($\mathrm{NFE}=2$) to generate 40-step action chunks on a 50\,Hz action grid. Approximate inference latency on an NVIDIA RTX 5090 GPU is 30--70\,ms per prediction, depending on model size.

\section{Hardware and Teleoperation}
\label{app:setup}
The robot has two fisheye cameras, one on each wrist, and a ZED Mini stereo camera on the head. Teleoperation uses a Manus Metaglove Pro and a Vive Tracker 3.0: the glove controls the fingers by retargeting the operator's hand skeleton to Wuji joint angles, and the tracker maps wrist motion to the arm's TCP pose.

\section{Experimental Results}
\label{app:eval}
Each policy is evaluated over 50 real-world trials: 24 nominal trials (8 training objects $\times$ 3) and 26 trials under conditions not covered by the real demonstrations: red lighting (8 objects $\times$ 1), a faster belt (8 objects $\times$ 1), three novel objects ($\times$ 3 each), and one trial combining all new conditions. Co-trained policies use 100 real demonstrations and 1,500 simulated trajectories. Table~\ref{tab:results-summary} aggregates successes by condition.
 
\begin{table}[h]
  \centering
  \footnotesize
  \caption{Successes per condition. Shifted = all non-nominal trials.}
  \label{tab:results-summary}
  \begin{tabular}{lccccccc}
    \toprule
    Policy & Nominal & Red light & Fast belt & Novel obj. & Combined & Shifted & Total \\
           & /24 & /8 & /8 & /9 & /1 & /26 & /50 \\
    \midrule
    \multicolumn{8}{l}{\textit{500M policy, trained from scratch}} \\
    Real 100 & 16 & 2 & 4 & 4 & 0 & 10 & 26 \\
    Fully grounded & 23 & 7 & 6 & 6 & 1 & 20 & 43 \\
    World grounded & 21 & 3 & 8 & 6 & 0 & 17 & 38 \\
    Behavior grounded & 20 & 3 & 6 & 4 & 1 & 14 & 34 \\
    Fully ungrounded & 18 & 0 & 6 & 5 & 0 & 11 & 29 \\
    \midrule
    \multicolumn{8}{l}{\textit{2B foundation model, post-trained}} \\
    Real 100 & 14 & 8 & 2 & 4 & 0 & 14 & 28 \\
    Real 1000 & 24 & 8 & 6 & 3 & 0 & 17 & 41 \\
    Fully grounded & 24 & 8 & 6 & 6 & 1 & 21 & 45 \\
    Fully ungrounded & 22 & 6 & 8 & 4 & 0 & 18 & 40 \\
    \midrule
    \multicolumn{8}{l}{\textit{$\pi_{0.5}$, fine-tuned}} \\
    Real 100 & 13 & 3 & 1 & 2 & 0 & 6 & 19 \\
    Real 1000 & 24 & 7 & 8 & 4 & 1 & 20 & 44 \\
    \bottomrule
  \end{tabular}
\end{table}

\end{appendices}

\end{document}